\documentclass[runningheads]{llncs}

\usepackage{eccv}

\usepackage{eccvabbrv}

\usepackage{graphicx}
\usepackage{booktabs}  

\usepackage[accsupp]{axessibility}  

\usepackage{hyperref}

\usepackage{orcidlink}

\usepackage{bbm}
\usepackage{multirow}   
\usepackage{colortbl}   
\usepackage[table,xcdraw]{xcolor} 
\usepackage{array}      
\usepackage{caption}    
\usepackage{amssymb}
\usepackage{tabularx}   
\usepackage{makecell}   
\usepackage{siunitx}
\usepackage{pifont}
\usepackage{tcolorbox}

\newcommand{\equalcontrib}{\textsuperscript{*}}
\newcommand{\corresponding}{\textsuperscript{\ensuremath{\dagger}}}

\begin{document}

\title{NutriBench-Kitchen: Benchmarking Embodied AI for Nutrition Management} 

\titlerunning{NutriBench-Kitchen}

\author{Yulin Wei\inst{1}\orcidlink{0009-0008-1594-0974}\equalcontrib \and
Xiangchen Wang\inst{2}\orcidlink{0009-0002-0857-4378}\equalcontrib \and
Jianhui Pan\inst{2} \and
Jinyu Xiao\inst{2} \and
Zheng Tan\inst{2} \and
Ruozai Tian\inst{2} \and
Guanhua Chen\inst{2}\orcidlink{0000-0002-5353-9734}\corresponding\and
Feng Zheng\inst{2,3}\orcidlink{0000-0002-1701-9141}\corresponding}

\authorrunning{Y.~Wei et al.}

\institute{Georgia Institute of Technology, Atlanta, USA \and
Southern University of Science and Technology, Shenzhen, China
\and
SpatialTemporal AI, Shenzhen, China
}
\begingroup
\renewcommand\thefootnote{}
\footnotetext[1]{* Equal contribution}
\footnotetext[2]{$\dagger$ Corresponding author}
\endgroup

\maketitle

\begin{abstract}

An embodied kitchen assistant must do more than recognize food in isolated frames. It must track ingredient states over time and integrate visual observations with recipe and nutritional knowledge to support constraint-aware decision-making. We formalize this capability as \emph{Embodied Nutrition Management}: perceiving nutrition-relevant events, maintaining a persistent food state, and using it for knowledge-grounded planning.
Existing benchmarks evaluate static food understanding or embodied cooking actions, but do not measure whether an agent can continuously update and use nutrition-relevant states in dynamic kitchens. To fill this gap, we introduce \textbf{NutriBench-Kitchen}, a benchmark containing 1,500 manually verified question--answer pairs from 160 cooking videos.
It covers five task families: Ingredient Entry, Memory Management, Recipe Query, Long-Term Planning, and Short-Term Planning, spanning food-state construction, maintenance, knowledge retrieval, and decision-making across different planning horizons.
Evaluations of proprietary and open-source large vision-language models reveal a substantial gap from human performance, particularly in quantitative ingredient estimation, long-term state tracking, and reasoning under interacting constraints.
We further introduce \textbf{Nutri-Vgent}, a diagnostic long-video agent with separate episodic, food-state, and recipe memories. Its consistent improvements demonstrate the value of explicit state representations and structured memory for nutrition management. Together, NutriBench-Kitchen and Nutri-Vgent provide a testbed for studying persistent state tracking and knowledge-grounded reasoning in dynamic kitchens.
Code is available at \url{https://github.com/V1ol1n/NutriBench-Kitchen}.

\keywords{Nutrition Management \and Egocentric Vision \and Embodied AI}
\end{abstract} 

\section{Introduction}

\begin{figure*}[t]
\centering
\includegraphics[width=1.0\linewidth]{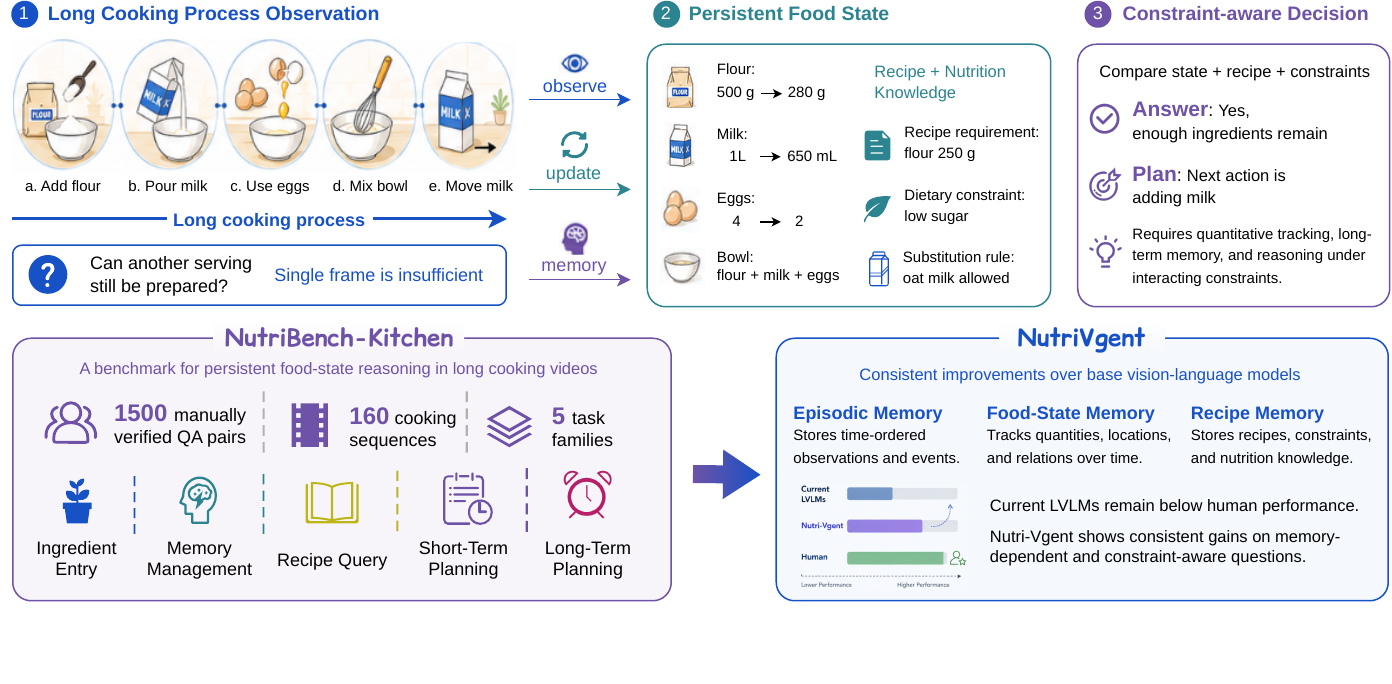}
\vspace{-5mm}
\caption{\textbf{Overview of Embodied Nutrition Management, NutriBench-Kitchen, and Nutri-Vgent.}
NutriBench-Kitchen evaluates five interconnected capabilities, from registering ingredients and maintaining persistent food states to querying recipes and planning at different temporal scales. Nutri-Vgent addresses these tasks using separate episodic, food-state, and recipe memories.}
\label{fig}
\end{figure*}

Consider an embodied assistant that has observed a cook preparing a dish for several minutes. Some flour has already been poured into a bowl, part of the milk has been used, and several ingredients are no longer visible on the counter. When asked whether another serving can still be prepared, the assistant cannot answer based on the current frame alone. It must recover earlier events, estimate the remaining ingredients, and compare the current inventory with the recipe requirements and specified dietary constraints. A correct answer therefore depends on a food state that has been accumulated and updated throughout the cooking process. Existing food benchmarks rarely evaluate this form of reasoning. They typically focus on ingredient recognition, nutrition estimation, or recipe understanding from static images, short observations, or structured records~\cite{farinella2014benchmark,marin2021recipe1m+,haussmann2019foodkg,ma2023food,thames2021nutrition5k,qi2025advancing,xu2025sfood,coburn2025comprehensive}. Egocentric datasets such as EPIC-KITCHENS~\cite{damen2018scaling,Damen2021PAMI}, Ego4D~\cite{grauman2022ego4d}, and HD-EPIC~\cite{perrett2025hd} bring the setting closer to real kitchens by providing long cooking activities and rich object interactions, but their evaluation usually targets action recognition, event localization, or temporal understanding rather than nutrition-relevant state updates.  Embodied-agent work~\cite{zhou2025mathcalp3versatileembodiedagents, luo2025visualembodiedbrainlet} further broadens the discussion to perception, planning, and control, while planning benchmarks~\cite{cai2025cookbench,zhang2025plan} often assume that the current environment state is already available. These settings leave a fundamental question unresolved: can a model integrate observations distributed over time into an updated food state and use that state to support a later decision?

We formulate this problem as \textbf{Embodied Nutrition Management}: the ability of an embodied agent to derive an evolving food state from continuous visual observations and use that state for recipe queries and constraint-aware planning. The central challenge is to preserve the consequences of earlier events after the relevant actions or objects are no longer visible. Observing flour being added to a bowl, for example, does not directly indicate whether enough remains for another batch. The agent must estimate the amount used, update the available inventory, and retain this change when answering a later question. Similar challenges arise when ingredients are divided among containers, partially consumed, substituted, or transformed during cooking. The resulting state may also depend on multiple information sources: visual evidence indicates what occurred, whereas recipe and nutritional knowledge determine whether the resulting inventory satisfies a request. Access to previous frames is useful, but insufficient when a model cannot convert observations into persistent state updates that remain available throughout the task.

We introduce \textbf{NutriBench-Kitchen}, a benchmark designed to evaluate this capability directly. It contains 1,500 manually verified question--answer pairs derived from 160 cooking sequences, including 145 HD-EPIC segments and 15 self-recorded long-form videos. The benchmark covers five task families: \textit{Ingredient Entry}, \textit{Memory Management}, \textit{Recipe Query}, \textit{Long-Term Planning}, and \textit{Short-Term Planning}. Together, these tasks span the process of registering ingredients, maintaining their changing states, retrieving recipe knowledge, and selecting actions across different planning horizons. HD-EPIC provides diverse, naturally occurring cooking activities, whereas the self-recorded videos provide directly measured ingredient quantities and nutritional values that cannot be reliably recovered from public footage. Each question is manually checked against the corresponding visual evidence to ensure correct grounding and annotation. The benchmark further includes disjoint Easy and Hard regimes that differ in whether hints to the relevant evidence are provided.

Questions in NutriBench-Kitchen cannot be answered by retrieving a single salient frame. A model may need to combine an early observation of the initial quantity of an ingredient with a later action that consumes part of it, and then compare the remaining amount with a recipe requirement introduced elsewhere in the dialogue. In other cases, the final state is never shown explicitly and must be inferred from multiple intermediate changes. Questions may also involve substitutions or dietary constraints that interact with the visually inferred state. This design distinguishes models that recognize isolated events from those that maintain a usable representation of the current kitchen state. It also reveals errors that conventional video question answering may overlook. A model may retrieve the frame in which an ingredient was added but still answer incorrectly because it fails to propagate the resulting inventory change.

Our experiments reveal a gulf between current large vision-language models and human performance. Errors are particularly common in quantitative ingredient estimation, long-range state maintenance, and reasoning over interactions between inventory conditions and dietary constraints. Models often recognize that an ingredient has been used but fail to incorporate this event into later answers. Increasing model scale does not consistently resolve these failures, and longer visual context does not uniformly improve performance. These results suggest that retrieving additional observations alone does not ensure that their consequences are preserved for subsequent reasoning. To examine this issue, we develop \textbf{Nutri-Vgent}, a diagnostic long-video agent equipped with structured memory. Episodic Memory stores temporally grounded observations, Food-State Memory maintains the current ingredient inventory, and Recipe Memory provides relevant recipe and nutritional knowledge. Nutri-Vgent consistently improves the performance of its underlying vision-language models, with the largest gains on questions involving earlier ingredient use, accumulated state changes, or multiple interacting constraints. These improvements indicate that explicit food-state tracking preserves information that is often lost during general long-context reasoning.

Our contributions are threefold:

\begin{itemize}
\item We formalize \textbf{Embodied Nutrition Management} as the joint problem of perceiving nutrition-relevant events, maintaining persistent food states, and performing knowledge-grounded planning in dynamic kitchens.

\item We introduce \textbf{NutriBench-Kitchen}, a benchmark of 1,500 manually verified question--answer pairs covering five balanced task families and two equally sized, disjoint evaluation regimes.

\item We systematically evaluate large vision-language models and introduce \textbf{Nutri-Vgent}, a structured-memory agent that demonstrates the value of explicit food-state representations for long-horizon, constraint-aware reasoning.

\end{itemize}
\section{Related Work}
\label{sec:related_work}

\subsection{Nutrition Management in Embodied AI}
Household embodied agents are increasingly expected to handle tasks that extend beyond atomic commands such as ``pick up the apple.'' In a kitchen, a useful assistant must track which ingredients are available, how they change during preparation, and whether the resulting state still satisfies a dietary or recipe constraint. Nutrition-aware AI has made progress on calorie estimation, ingredient recognition, and cross-modal recipe retrieval, as represented by Nutrition5K~\cite{thames2021nutrition5k}, Recipe1M+~\cite{marin2021recipe1m+} and FoodKG~\cite{haussmann2019foodkg}. These resources, however, usually assume a static input or database state and therefore do not capture everyday cooking changes such as partial use, cutting, transfer, or depletion.
Embodied and egocentric vision datasets~\cite{grauman2022ego4d,damen2018scaling} provide long observations of manipulation, object interaction, and human activity. Their evaluation targets are often action recognition, localization, or procedural understanding. The missing link is how visual events update nutrition-relevant kitchen states: what changed, what remains, and what decision is still feasible.

\begin{table*}[t]
\centering
\setlength{\tabcolsep}{4.6pt}
\renewcommand{\arraystretch}{1.15}
\caption{Coverage of capabilities required for embodied nutrition management. A check mark indicates that the capability is an explicit task, annotation target, or evaluation objective of the benchmark/resource, rather than a capability that could only be inferred from the raw data. Vis.: visual food or event grounding; Ing./Nutri.: ingredient, quantity, or nutrition estimation; State: temporal food-state or inventory tracking; Food Know.: recipe, cuisine, or food-knowledge reasoning; Constraint: inventory- or diet-constrained decision/query; Plan.: long- or short-horizon procedural planning.}
\label{tab:bench-capability}
\resizebox{\textwidth}{!}{%
\begin{tabular}{lcccccc}
\toprule
\textbf{Benchmark / Resource} & \textbf{Vis.} & \textbf{Ing./Nutri.} & \textbf{State} & \textbf{Food Know.} & \textbf{Constraint} & \textbf{Plan.} \\
\midrule
UNICT-FD889~\cite{farinella2014benchmark} & \ding{51} & \ding{55} & \ding{55} & \ding{55} & \ding{55} & \ding{55} \\
Food-500 Cap~\cite{ma2023food} & \ding{51} & \ding{51} & \ding{55} & \ding{55} & \ding{55} & \ding{55} \\
Recipe1M+~\cite{marin2021recipe1m+} & \ding{51} & \ding{51} & \ding{55} & \ding{51} & \ding{55} & \ding{55} \\
Nutrition5k~\cite{thames2021nutrition5k} & \ding{51} & \ding{51} & \ding{55} & \ding{55} & \ding{55} & \ding{55} \\
FastFood~\cite{qi2025advancing} & \ding{51} & \ding{51} & \ding{55} & \ding{55} & \ding{55} & \ding{55} \\
SFOOD~\cite{xu2025sfood} & \ding{51} & \ding{51} & \ding{55} & \ding{55} & \ding{55} & \ding{55} \\
FoodieQA~\cite{li2024foodieqa} & \ding{51} & \ding{55} & \ding{55} & \ding{51} & \ding{55} & \ding{55} \\
WorldCuisines~\cite{winata2024worldcuisines} & \ding{51} & \ding{55} & \ding{55} & \ding{51} & \ding{55} & \ding{55} \\
HD-EPIC~\cite{perrett2025hd} & \ding{51} & \ding{51} & \ding{55} & \ding{51} & \ding{55} & \ding{55} \\
ET-Plan-Bench~\cite{zhang2025plan} & \ding{55} & \ding{55} & \ding{55} & \ding{55} & \ding{55} & \ding{51} \\
CookBench~\cite{cai2025cookbench} & \ding{51} & \ding{55} & \ding{55} & \ding{51} & \ding{55} & \ding{51} \\
\rowcolor{gray!10} \textbf{NutriBench-Kitchen (ours)} & \textbf{\ding{51}} & \textbf{\ding{51}} & \textbf{\ding{51}} & \textbf{\ding{51}} & \textbf{\ding{51}} & \textbf{\ding{51}} \\
\bottomrule
\end{tabular}%
}
\end{table*}

\subsection{Evaluation and Benchmarks}
Existing benchmarks usually evaluate only part of this problem. Vision-centric resources such as SFOOD~\cite{xu2025sfood} and HD-EPIC~\cite{perrett2025hd} provide rich visual signals for food attributes, ingredient classification, or egocentric kitchen activity, but they do not require models to carry those observations forward into nutrition- or recipe-constrained decisions. Text-centric and QA benchmarks such as FoodieQA~\cite{li2024foodieqa} and WorldCuisines~\cite{winata2024worldcuisines} evaluate food knowledge, but lack the embodied dimension of updating that knowledge from a changing kitchen scene. Planning benchmarks such as CookBench~\cite{cai2025cookbench} and ET-Plan-Bench~\cite{zhang2025plan} evaluate procedural decision-making, while generally leaving nutrition constraints and persistent ingredient-state changes outside the task.
As summarized in Table~\ref{tab:bench-capability}, existing resources usually cover one or two parts of the nutrition-management lifecycle rather than the full loop from video-grounded food-state updates to recipe, nutrition, and planning decisions under changing inventory constraints. \textbf{NutriBench-Kitchen} fills this gap; the concrete task formulation and capability definitions are introduced in \cref{sec:task}.

\section{Embodied Nutrition Management}
\label{sec:task}

\subsection{Task Scope and Formulation}
\label{sec:task_def}

We define \textbf{Embodied Nutrition Management} as the ability to infer, maintain, and use nutrition-relevant kitchen states from long-form cooking observations. Let $V=\{v_t\}_{t=1}^{T}$ denote a cooking video, $q$ a question, $h$ an optional dialogue history, and $r$ an optional textual frame hint. A model predicts an answer
\begin{equation}
    \hat{y}=F(V,q,h,r;\mathcal{K}),
\end{equation}
where $\mathcal{K}$ denotes permitted external recipe or nutrition knowledge. The benchmark evaluates whether the model can recover a latent kitchen state $S_t$ from temporally distributed visual evidence and use that state to answer questions about ingredients, recipes, nutritional constraints, and feasible future steps.

This formulation is intentionally observational. The model does not change the environment, receive simulator feedback, or execute physical actions. Instead, NutriBench-Kitchen evaluates the perception, memory, and reasoning prerequisites that an embodied assistant would need before acting in the physical world. In particular, Long-Term Planning and Short-Term Planning are treated as plan understanding and decision reasoning tasks: a model must infer what stage has been reached, what dependencies have been satisfied, or what immediate next decision is appropriate, but it is not asked to perform closed-loop control.

The core difficulty is that the answer often depends on state changes that are no longer visible. A model may need to recognize that an ingredient was added several minutes earlier, estimate how much was consumed, retain the consequence after the ingredient disappears from view, and compare the resulting state with a recipe or dietary constraint. This requires more than retrieving a salient frame. The model must convert visual events into persistent food-state updates and ground later reasoning in those updates.

\subsection{NutriBench-Kitchen Benchmark Construction}
\label{sec:benchmark_construction}

\begin{figure*}[t]
    \centering
    \includegraphics[width=1.0\linewidth]{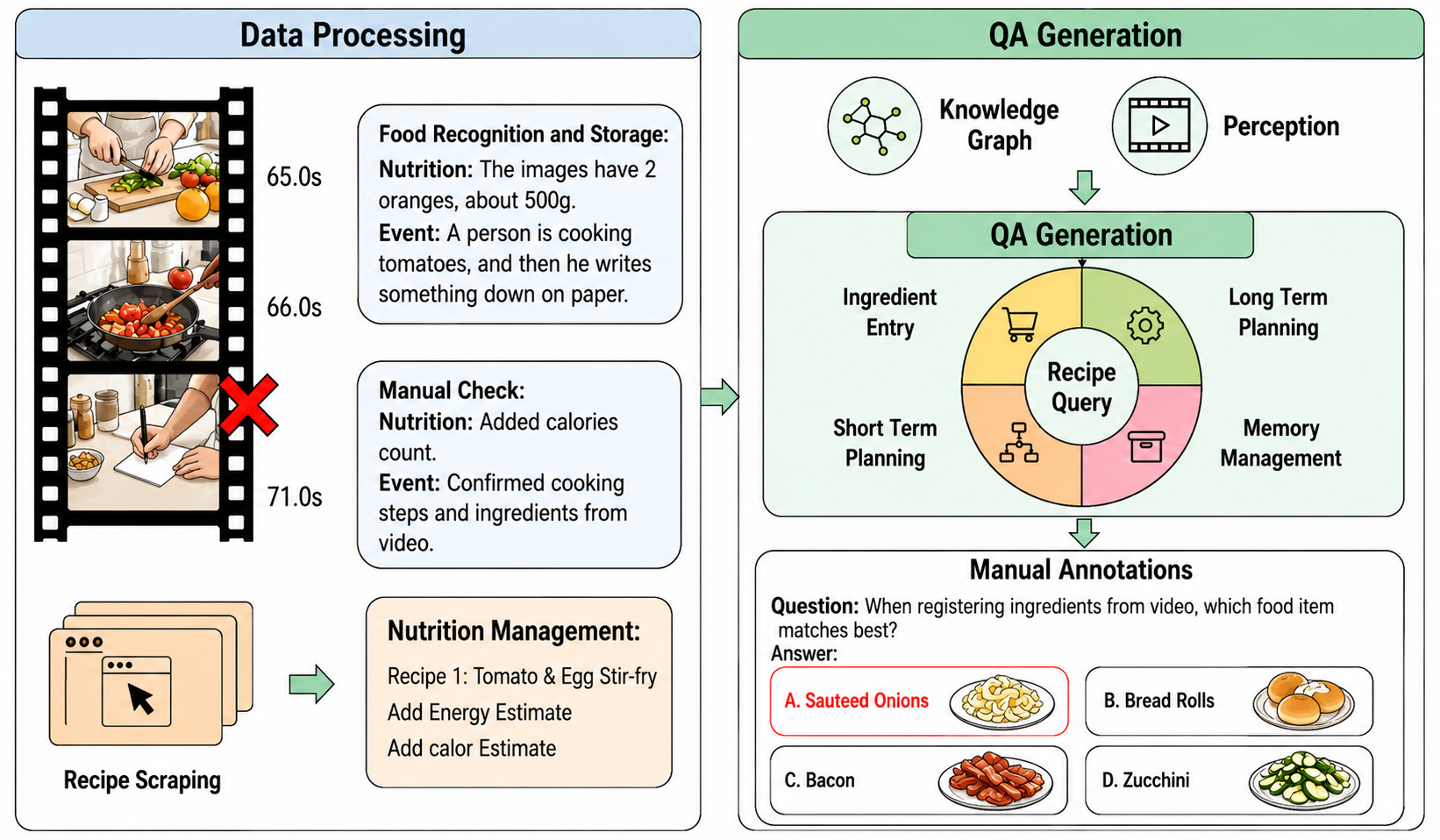}
    \vspace{-5mm}
    \caption{The construction pipeline of NutriBench-Kitchen. The benchmark is built from long-form cooking videos and recipe resources through video preprocessing, perception-assisted food recognition, recipe and nutrition knowledge construction, task-oriented QA generation, and final manual annotation. Model outputs are used only as annotation drafts; final food states, evidence spans, answers, and distractors are human verified.}
    \label{fig:data-pipeline}
\end{figure*}

To evaluate Embodied Nutrition Management, we construct \textbf{NutriBench-Kitchen}, a benchmark of 1,500 manually verified question--answer pairs from 160 long-form cooking videos. The videos include 145 selected HD-EPIC cooking segments~\cite{perrett2025hd} and 15 self-recorded long-form cooking videos. These two sources provide complementary supervision: HD-EPIC contributes diverse, naturally occurring egocentric kitchen activity, while the self-recorded subset provides controlled ingredient-state changes with directly measured quantities and nutritional values. The overall construction process follows the data-processing and QA-generation workflow shown in \cref{fig:data-pipeline}.

\paragraph{Data processing.}
We first curate cooking videos that contain meaningful ingredient interactions, visible preparation steps, and temporal state changes. Long videos are temporally processed into sampled visual evidence, with denser coverage around action-rich intervals such as adding, transferring, cutting, mixing, consuming, or discarding ingredients. Near-duplicate frames are removed with perceptual hashing to reduce redundant evidence. When audio or transcript information is available, it is aligned with the visual stream as auxiliary context for annotation, while visual evidence remains the primary grounding source for benchmark questions.

\paragraph{Recipe and nutrition knowledge construction.}
In parallel, we collect recipe resources and construct a Recipe Knowledge Graph from more than 10,000 recipes. The graph links recipes, ingredients, tools, procedural steps, nutritional attributes, and candidate substitutions. Ingredient aliases are canonicalized so that visually observed foods can be matched to recipe-level concepts. Nutrition information is associated with verified ingredient identities and quantities, enabling questions about calorie ranges, ingredient availability, dietary restrictions, and substitutions. The graph supports knowledge-grounded QA generation, but automatically collected recipe relations are treated as structured resources rather than unverified ground truth for visual events.

\paragraph{Perception-assisted annotation.}
We use Qwen3-VL-max~\cite{qwen3_vl} to generate initial open-vocabulary proposals for food identities, nutrition-relevant events, and candidate temporal evidence. These outputs serve only as drafts. Human annotators inspect the original video frames, correct ingredient identities and states, align relevant temporal spans, and resolve visually ambiguous cases. For the self-recorded videos, ingredient weights are physically measured during recording and used to compute nutritional values. For HD-EPIC segments, annotators estimate quantities from the available visual evidence and reconcile disagreements through consensus. This source-dependent protocol preserves both controlled quantitative supervision and in-the-wild visual diversity.

\paragraph{QA generation and manual verification.}
QA candidates are generated from task-specific templates that combine perception evidence, current food states, and recipe-knowledge constraints. The QA generation module is organized around five task families: Ingredient Entry, Memory Management, Recipe Query, Long-Term Planning, and Short-Term Planning. Candidate questions are filtered by LLM self-consistency and then manually verified. Annotators check whether each question is visually grounded, whether the answer follows from the cited evidence and knowledge constraints, whether distractors are plausible but incorrect, and whether the assigned task label matches the required reasoning process. The final benchmark is balanced across the five task families and supports two evaluation regimes described below.

\subsection{Task Families and Evaluation Regimes}
\label{sec:task_families}

\begin{figure*}[t]
    \centering
    \includegraphics[width=1.0\linewidth]{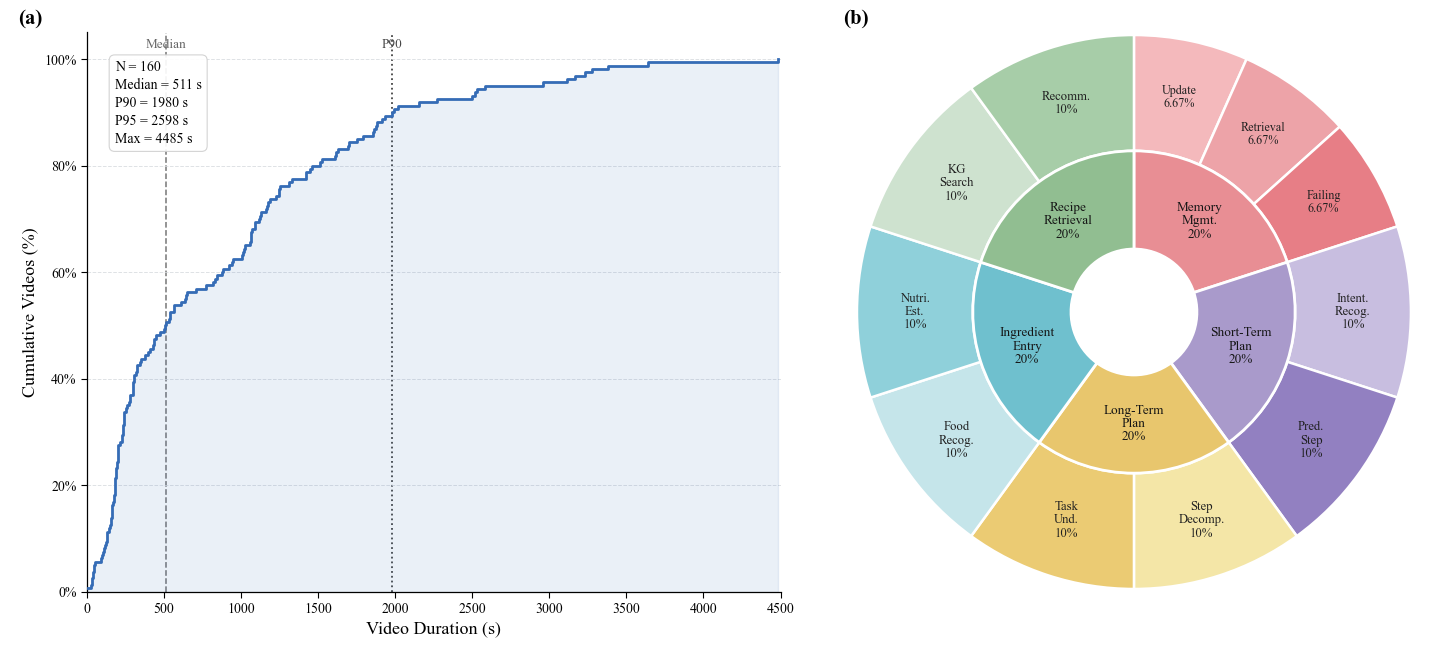}
    \vspace{-5mm}
    \caption{Statistics of NutriBench-Kitchen. \textbf{Left}: video duration distribution. The benchmark covers a wide temporal range, including long cooking episodes that require extended memory. \textbf{Right}: task taxonomy distribution. NutriBench-Kitchen is balanced across five task families and further divided into fine-grained subcategories for perception, state tracking, recipe reasoning, and planning-oriented decision understanding.}
    \label{fig:Visual}
\end{figure*}

NutriBench-Kitchen evaluates five interconnected task families, corresponding to the QA-generation module in \cref{fig:data-pipeline}. Together, they cover the process of registering ingredients, maintaining food states, querying recipe knowledge, and reasoning about future steps at different temporal scales.

\paragraph{Ingredient Entry.}
Ingredient Entry evaluates whether a model can identify a relevant ingredient and estimate its associated quantity or nutritional value from the observed cooking process. Questions may require integrating multiple views or time points because an ingredient can be partially occluded, transferred between containers, or visually ambiguous after mixing. This task primarily probes fine-grained visual grounding and quantitative estimation.

\paragraph{Memory Management.}
Memory Management evaluates whether a model can reconstruct the current ingredient or inventory state after a sequence of events. The answer may depend on additions, consumption, relocation, disposal, or transformations observed earlier in the video. Formally, the model must infer an updated state $S_t$ from the previous state and the relevant observations in $V_{1:t}$, even though $S_t$ is not directly provided.

\paragraph{Recipe Query.}
Recipe Query evaluates whether a model can select or assess a recipe under constraints derived from the observed inventory and structured nutritional knowledge. Questions may involve ingredient availability, substitutions, calorie ranges, or dietary restrictions. Success therefore requires both knowledge retrieval and consistency with the current video-grounded state.

\paragraph{Long-Term Planning.}
Long-Term Planning evaluates high-level procedural reasoning over an extended recipe or cooking episode. Questions test ordering, dependency, completion, and decomposition of semantically meaningful stages. The task does not require a model to execute the plan; it measures whether the model understands the long-range structure that would be needed for execution.

\paragraph{Short-Term Planning.}
Short-Term Planning evaluates the immediate next decision given the current observation, active subgoal, and inferred kitchen state. Questions are grounded in local object interactions but can still depend on earlier events, for example when the validity of the next step depends on whether an ingredient has already been added or a preparation stage has been completed.

Each task family is evaluated under Easy and Hard regimes. The \textbf{Easy} regime contains single-turn questions accompanied by a textual \texttt{frame\_hint} that summarizes relevant local visual context. The \textbf{Hard} regime contains different, more complex questions, uses multi-turn dialogue context, and does not provide the frame hint. Easy and Hard are therefore not paired transformations of the same questions. Their scores characterize two benchmark regimes: Easy measures reasoning with localized textual support, while Hard places greater pressure on independent visual perception, long-range memory, and dialogue-grounded state tracking.

\section{Nutri-Vgent}

Traditional large vision-language models (VLMs) face a critical bottleneck in embodied environments: the inability to bridge continuous visual perception with long-horizon dietary reasoning. While recent visual retrieval-augmented generation (RAG) frameworks, such as Vgent, have made strides in long-video understanding, they primarily rely on generic, unstructured memory retrieval. This flat retrieval approach falls short in complex realistic kitchen scenarios, which strictly demand precise inventory tracking and goal-oriented planning. To overcome this limitation, we propose \textbf{Nutri-Vgent}, which significantly advances the base Vgent architecture by introducing a novel, domain-specific structured memory tailored for dietary management. Rather than treating past observations as a homogeneous pool, our agent explicitly decouples and continuously updates three core components: episodic experience, persistent food states, and recipe constraints. This specialized design tightly couples the memory mechanism with kitchen dynamics, enabling robust, constraint-aware reasoning over extended periods.

\noindent \textbf{Nutrition Memory construction.}
The core design choice is to separate transient visual evidence from the persistent food state that later questions should use, echoing recent findings that multimodal models need explicit mechanisms for remembering and recalling spatially grounded evidence~\cite{yang2024thinking}. Episodic Memory stores temporally grounded observations, including actions, object states, locations, and relations. Food Memory maintains the current ingredient inventory, with fields such as canonical identity, estimated quantity, freshness, location, and timestamp. Recipe Memory stores external recipe knowledge normalized into ingredient, step, and tool relations. These three memories play different roles: Episodic Memory explains how a state was reached, Food Memory represents what is currently available, and Recipe Memory defines what constraints or substitutions are valid.

Nutri-Vgent constructs these memories from the video stream with a sliding temporal window. For each segment, the backbone LVLM summarizes nutrition-relevant events and writes them to Episodic Memory with temporal anchors. In parallel, food entities are canonicalized against the recipe vocabulary and used to update the Food Memory ledger. Actions such as adding, cutting, transferring, consuming, or discarding an ingredient are treated as state-changing events rather than isolated captions, so the inventory can be updated even after the visual evidence is no longer present. This structured update process is the main difference from a generic retrieved-context pool: the memory does not only say what was observed, but also records how the observed event changes the current food state.

\begin{figure*}[t]
    \centering
    \includegraphics[width=1.0\linewidth]{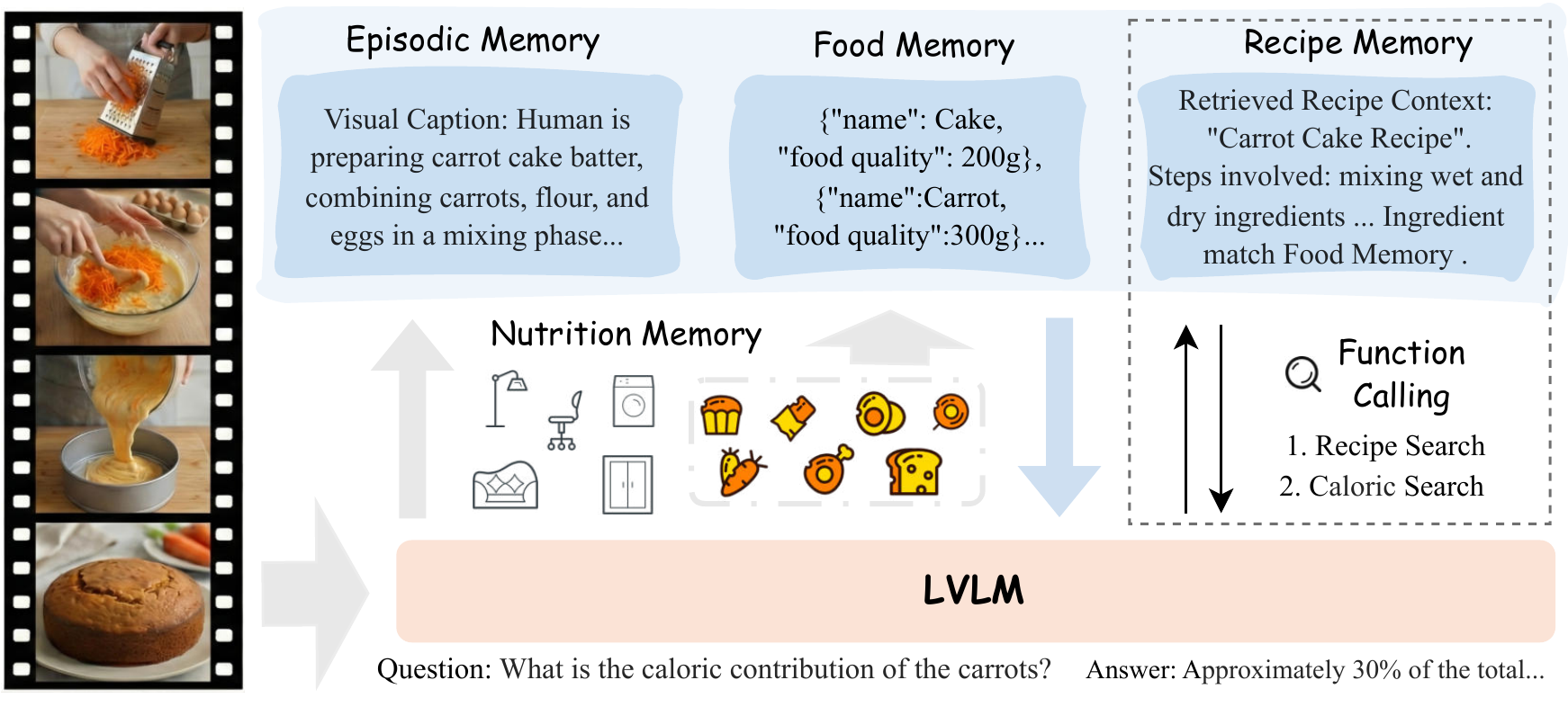} 
    \vspace{-5mm}
    \caption{\textbf{Overview of the Nutri-Vgent architecture.} Nutri-Vgent processes continuous video streams with Nutrition Memory, which separates information into three persistent modules: Episodic Memory, Food Memory, and Recipe Memory. Structured memory and function calling support constraint-aware nutritional reasoning with the backbone LVLM.}
    \label{fig:agent-pipeline}
\end{figure*}

\noindent \textbf{Memory-grounded reasoning.} 
At inference time, the user question is interpreted against the current Nutrition Memory. Episodic Memory provides evidence for recent and long-range events, while Food Memory supplies the active inventory constraints, including quantities, locations, and freshness states. When the question requires recipe or caloric knowledge, Nutri-Vgent accesses Recipe Memory through prompt-level function calls. DishSpec retrieves the canonical ingredients and reference gram weights for a requested dish, and FoodByCalories retrieves candidate foods or substitutions that satisfy a requested calorie range. The returned candidates are checked against the Food Memory ledger before the LVLM produces the answer. The same architecture therefore supports both state-estimation questions and recipe-selection questions: visual evidence establishes the evolving kitchen state, recipe memory defines the relevant constraints, and the final response is generated only after these two sources are reconciled.

\section{Experiment}
\label{sec:experiment}

In this section, we evaluate the performance of various baselines and our Nutri-Vgent on {NutriBench-Kitchen}. We aim to answer three key questions: (1) How challenging is the proposed benchmark for current state-of-the-art multimodal models? (2) To what extent does our proposed hybrid memory architecture enhance the agent's capability in long-horizon nutrition management? (3) What are the remaining challenges and future directions for this domain?

\begin{table*}[t]
\centering
\setlength{\tabcolsep}{2.9pt}
\renewcommand{\arraystretch}{1.12}
\caption{Benchmark results reorganized into five task categories, each with Easy(\textbf{E.}) and Hard (\textbf{H.}).  IE for Ingredient Entry. Instruction Decomposition, Instruction Understanding, Inventory Management and Recipe Query. }
{\small
\begin{tabular}{l c *{10}{c}}
\toprule
\multirow{2}{*}{\textbf{Models}} & 
\multicolumn{2}{c}{\textbf{IE}} &
\multicolumn{2}{c}{\textbf{LTP}} &
\multicolumn{2}{c}{\textbf{MM}} &
\multicolumn{2}{c}{\textbf{RQ}} &
\multicolumn{2}{c}{\textbf{STP}} & \multirow{2}{*}{\textbf{Avg.}}\\
& \textbf{E.} & \textbf{H.} & \textbf{E.} & \textbf{H.}
& \textbf{E.} & \textbf{H.} & \textbf{E.} & \textbf{H.} & \textbf{E.} & \textbf{H.} & \\
\midrule
Chance (Random)  & 16.7 & 25.0 & 25.0 & 25.0 & 25.0 & 25.0 & 25.0 & 25.0 & 25.0 & 25.0 & 24.2\\
Chance (Frequency)  & 12.6 & 6.2 & 28.5 & 22.4 & 5.3 & 2.1 & 29.8 & 23.5 & 34.2 & 26.8 & 19.1\\
Human  & 92.5 & 88.0 & 95.0 & 91.5 & 89.0 & 84.5 & 96.0 & 92.0 & 97.5 & 94.0 & 92.0 \\
\midrule
\multicolumn{12}{c}{\textit{Proprietary LVLMs}} \\
Gemini 2.5 Pro ~\cite{comanici2025gemini25pushingfrontier} & 46.7 & 14.7 & 95.3 & 78.0 & 47.3 & 66.7 & 65.3 & 69.3 & 64.7 & 63.3 & 61.1 \\
GPT-4o ~\cite{openai_gpt4o}  & 30.7 & 16.7 & 69.3 & 66.7 & 26.7 & 61.3 & 46.7 & 50.5 & 49.3 & 59.3 & 47.7 \\
\midrule
\multicolumn{12}{c}{\textit{Open-Source LVLMs}} \\

RoboBrain2.0-7B~\cite{baairobobrainteam2025robobrain20technicalreport}  & 29.3 & 2.0 & 86.0 & 58.0 & 24.7 & 52.0 & 53.3 & 48.5 & 53.3 & 48.0 & 45.5 \\
VeBrain-7B~\cite{luo2025visualembodiedbrainlet} & 34.7 & 2.7 & 82.7 & 57.3 & 40.7 & 68.7 & 67.3 & 66.3 & 52.0 & 53.3 & 52.6 \\
InternVL3.5-4B-HF~\cite{wang2025internvl3}  & 30.7 & 33.3 & 70.7 & 45.3 & 32.7 & 54.0 & 54.0 & 50.5 & 48.0 & 46.7 & 46.6 \\
InternVL3.5-8B-HF~\cite{wang2025internvl3} & 26.7 & 18.0 & 81.3 & 45.3 & 36.7 & 58.7 & 50.7 & 53.5 & 48.7 & 49.3 & 46.9 \\
GLM-4.1V-9B-Base~\cite{hong2025glm}  & 38.0 & 5.3 & 90.7 & 57.3 & 34.7 & 64.0 & 58.0 & 60.4 & 50.0 & 50.0 & 50.8 \\
GLM-4.1V-9B-Think~\cite{hong2025glm}  & 16.7 & 4.0 & 28.0 & 22.7 & 20.7 & 41.3 & 52.7 & 33.7 & 37.3 & 37.3 & 29.4 \\
Qwen2.5-VL-3B~\cite{qwen2.5vl}  & 22.0 & 0.7 & 84.0 & 73.3 & 39.3 & 68.0 & 59.3 & 62.4 & 50.0 & 48.0 & 50.7 \\
Qwen2.5-VL-7B~\cite{qwen2.5vl}  & 30.7 & 3.3 & 84.7 & 52.0 & 34.0 & 63.3 & 62.0 & 60.4 & 59.3 & 58.7 & 50.8 \\
Qwen3-VL-4B~\cite{qwen3_vl} &  32.7 & 16.7 & 93.3 & 40.0 & 44.0 & 70.7 & 70.0 & 69.3 & 60.7 & 57.3 & 55.5 \\
\ \ + Vgent~\cite{shen2025vgentgraphbasedretrievalreasoningaugmentedgeneration} & 18.0 & 15.3 & 83.6 & 34.6 & 21.5 & 14.6 & 21.7 & 27.7 & 33.3 & 17.8 & 28.8 \\
\rowcolor{gray!10}\ \ + Nutri-Vgent & 22.7 & 73.3 & 72.7 & 35.1 & 42.7 & 68.7 & 61.3 & 95.0 & 76.0 & 73.0 & 62.1 \\
Qwen3-VL-8B~\cite{qwen3_vl} & 30.7 & 14.0 & 95.3 & 48.7 & 38.7 & 66.0 & 70.7 & 63.4 & 53.3 & 52.7 & 53.4 \\
\ \ + Vgent~\cite{shen2025vgentgraphbasedretrievalreasoningaugmentedgeneration} & 16.9 & 20.7 & 77.6 & 58.8 & 20.8 & 26.2 & 25.5 & 21.3 & 30.4 & 18.5 & 31.7 \\
\rowcolor{gray!10}\ \ + Nutri-Vgent & 20.0 & 73.3 & 68.0 & 53.5 & 42.0 & 69.3 & 58.0 & 94.1 & 73.3 & 71.3 & 62.3 \\
\bottomrule
\end{tabular}
}
\label{tab:recipe-tasks}
\end{table*}


\subsection{Implementation Details}
We evaluate proprietary models, including GPT-4o~\cite{openai_gpt4o} and Gemini 2.5 Pro~\cite{gemini_pro}, as well as open-source MLLMs from the InternVL3.5~\cite{internvl}, GLM-4V~\cite{hong2025glm}, and Qwen-VL~\cite{qwen3_vl} families. Nutri-Vgent is instantiated with Qwen3-VL-4B and Qwen3-VL-8B backbones. It employs a structured-memory architecture based on visual retrieval-augmented generation. We construct an offline video graph from uniformly sampled clips and dynamically retrieve relevant observations during inference to update Episodic Memory and Food-State Memory. All baseline models are evaluated zero-shot with the same chain-of-thought prompting protocol. Additional details, including retrieval thresholds, generation settings, and hardware configurations, are provided in the supplementary material.

Under this unified evaluation protocol, we use metrics tailored to the two answer formats in NutriBench-Kitchen. For multiple-choice questions, we report accuracy, with the predicted option extracted from each generated response using a deterministic parser. For numerical questions, which primarily occur in \textit{Ingredient Entry} and \textit{Memory Management}, we report Mean Relative Accuracy (MRA). Given $N$ examples and tolerance levels $\mathcal{C}={0.50,0.55,\ldots,0.95}$, MRA is defined as

\begin{equation}
\mathrm{MRA}
=
\frac{1}{N|\mathcal{C}|}
\sum_{i=1}^{N}
\sum_{\theta\in\mathcal{C}}
\mathbbm{1}
\left(
\frac{|\hat{y}_i-y_i|}{\max(|y_i|,\epsilon)}
<
1-\theta
\right),
\end{equation}

where $\epsilon$ prevents division by zero. MRA measures the proportion of predictions that satisfy progressively stricter relative-error tolerances, assigning higher scores to more accurate numerical estimates. Unless otherwise specified, task-level results are reported separately for the Easy and Hard regimes, and the overall score is computed as the unweighted mean across all task--regime cells.

\subsection{Main Results}
\label{sec:main_results}
~\cref{tab:recipe-tasks} presents the comparative results across five task dimensions. We summarize the key observations as follows:
\noindent \textbf{Benchmark Rigor and Challenge.} 
The benchmark presents a significant challenge to current AI systems. There is a substantial gap between human performance ($\sim92.0\%$ avg) and chance baselines (Random $\sim13.4\%$, Frequency $\sim19.1\%$). Crucially, while frequency-based guessing works reasonably well on ``Easy'' splits, it collapses on ``Hard'' splits, validating that our dataset relies on deep semantic understanding rather than statistical biases.

\noindent \textbf{Limitations of General-Purpose MLLMs.}
Despite strong pretraining, general-purpose MLLMs struggle with the core requirement of \emph{persistent} inventory tracking and constraint-grounded reasoning. A prominent failure mode is \emph{context forgetting} in long-horizon settings: for example, Qwen2.5-VL-7B drops sharply on \textit{Ingredient Entry (IE)} from 30.7\% (Easy) to 3.3\% (Hard), suggesting that without explicit mechanisms to retain and update state, models cannot reliably accumulate evidence across extended cooking procedures. Importantly, larger scale alone does not resolve this issue. Even {GPT-4o} achieves only 47.7\% on average and performs particularly poorly on \textit{Memory Management (MM)} (26.7\% on Easy), implying that generic reasoning capacity is insufficient when success hinges on precise, temporally consistent state updates rather than isolated perception or free-form explanation.Notably, Nutri-Vgent (4B) outperforms the much larger GLM-4V-9B (Avg 62.1\% vs 50.8\%) and Gemini 2.5 Pro (62.1\% vs 61.1\%). This indicates that for domain-specific embodied tasks, investing in structured reasoning frameworks is often more efficient than solely scaling up model parameters.

\begin{table*}[th]
  \centering
  \setlength{\tabcolsep}{3.5pt}
  \renewcommand{\arraystretch}{1.15}
  \caption{Ablation study on memory components using Qwen3-VL backbones. We isolate the effects of purely visual input (Vis.), Episodic Memory (Epi.), Food Memory (Food), and full structured memory (Epi. + Food) across five reasoning tasks: Ingredient Entry (IE), Long-Term Planning (LTP), Multimodal Matching (MM), Recipe Query (RQ), and Short-Term Planning (STP).}
  \label{tab:model_perf}
  \begin{tabular}{l| S S S| S S S S S S}
    \toprule
    \multirow{2}{*}{\textbf{Models}} & \multicolumn{3}{c|}{\textbf{Memory}} &
\multicolumn{6}{c}{\textbf{Performance}}  \\
 & \textbf{Vis.} & \textbf{Epi.} & \textbf{Food} & {\textbf{IE}} & {\textbf{LTP}} & {\textbf{MM}} & {\textbf{RQ}} & {\textbf{STP}} & {\textbf{Overall}} \\
    \midrule
    \multirow{4}{*}{Qwen3-VL-4B} & \ding{51} &  & & 24.70 & 66.65 & 57.35 & 69.65 & 59.00 & 55.5 \\
     &  & \ding{51} & & 50.00 & 44.30 & 46.30 & 44.55 & 51.00 & 47.3 \\
      &   &   & \ding{51} & 47.35 & 57.70 & 43.00 & 26.20 & 53.00 & 45.5 \\
    &  &  \ding{51} & \ding{51}  & 48.00 & 53.90 & 55.70 & 78.15 & 74.50 & 62.1 \\
    \midrule
    \multirow{4}{*}{Qwen3-VL-8B}  & \ding{51} &   &  & 22.35 & 72.00 & 52.35 & 67.05 & 53.00 & 53.4 \\
     &   &   \ding{51} & & 57.00 & 53.65 & 44.30 & 50.05 & 53.70 & 51.8 \\
     &   &  &   \ding{51} & 43.65 & 69.00 & 42.65 & 26.85 & 52.35 & 46.9 \\
    &  &   \ding{51} & \ding{51} & 46.65 & 60.75 & 55.65 & 76.05 & 72.30 & 62.3 \\
    \bottomrule
  \end{tabular}
\end{table*}


\noindent \textbf{Superiority of Nutri-Vgent.} 
Equipped with our novel structured memory architecture, \textbf{Nutri-Vgent} delivers transformative performance gains, effectively bridging the perception-reasoning gap.
\begin{itemize}
    \item \textit{Beating Giants with Smaller Models}: Nutri-Vgent (Qwen3-VL-8B) achieves an average score of 62.3\%, outperforming its base model (53.4\%), the larger GLM-4V-9B (50.8\%), and the proprietary GPT-4o (47.7\%). It even matches Gemini 2.5 Pro (61.1\%), proving that domain-specific structured architecture is more effective than pure parameter scaling.
    
    \item \textit{Structured Memory vs. Generic RAG}: While a generic RAG pipeline (Vgent) slightly improves specific tasks via basic episodic retrieval, its unstructured text concatenation often overwhelms the model, causing overall accuracy drops. Conversely, Nutri-Vgent explicitly decouples \textit{Food} and \textit{Recipe Memory} to accurately track dynamic states, effectively preventing context pollution and hallucinations.
    
    \item \textit{Resilience in Hard Scenarios}: Nutri-Vgent demonstrates striking robustness in complex, multi-turn splits. In \textit{Recipe Query (RQ)}, it surges from 63.4\% to 94.1\%. In \textit{Ingredient Entry (IE)}, it jumps from 14.0\% to 73.3\% (vastly exceeding standard Vgent's 24.7\% peak), proving its ability to reliably track ingredients that vanish from the visual field.
\end{itemize}

\noindent \textbf{Trade-offs and Analysis.} 
We observe a minor performance dip in some ``Easy'' tasks (e.g., IE Easy drops from 30.7\% to 20.0\%). We attribute this to \textit{retrieval overhead}: in simple, fully visible scenes, direct LVLM perception is sufficient. Enforcing strict Knowledge Graph consistency checks can occasionally "over-correct" these straightforward cases. Nevertheless, the massive gains achieved in complex, constraint-heavy scenarios overwhelmingly justify this architectural trade-off for real-world deployments.

\subsection{Ablation Studies}
\label{sec:ablation}

To rigorously evaluate the contribution of each component within our proposed architecture, we conducted an ablation study on the Qwen3-VL (4B and 8B) backbones, as shown in \cref{tab:model_perf}. We systematically isolated the effects of purely visual input (the base VLM), isolated Episodic Memory, isolated Food Memory, and our full structured memory approach (Episodic + Food).

\noindent \textbf{The Limitation of Pure Visual Input.} 
When relying solely on the \textit{Visual} modality without external memory, the models exhibit severe temporal forgetting. For instance, the visual-only Qwen3-VL-8B achieves a mere 22.35\% on Ingredient Entry (IE). This confirms that standard monolithic VLMs, regardless of their visual reasoning strength, are fundamentally unequipped to track fine-grained state changes over long videos.

\noindent \textbf{Single Memory Modules.} 
Introducing either \textit{Episodic Memory} or \textit{Food Memory} in isolation drastically improves state-tracking capabilities. The 8B model's IE performance jumps to 57.00\% (Episodic only) and 43.65\% (Food only). However, this comes at the cost of global reasoning: without the complementary visual context or a complete state representation, the average overall performance actually drops (from 53.4\% to 51.8\% and 46.9\% for the 8B model). Tasks requiring comprehensive situational awareness, such as Long-Term Planning (LTP) and Recipe Query (RQ), degrade significantly when the model is forced to rely on a single memory type.

\noindent \textbf{The Effectiveness of Hybrid Structured Memory.} 
The combination of \textit{Episodic Memory} and \textit{Food Memory} (our full Nutri-Vgent architecture) resolves this trade-off, achieving the highest overall scores (62.1\% for 4B and 62.3\% for 8B). This structured decoupling allows the agent to simultaneously track what just happened (Episodic) and what is currently available (Food) without context pollution. The synergy is most evident in complex downstream tasks: compared to the visual-only baseline, the 8B hybrid model boosts Recipe Query (RQ) from 67.05\% to 76.05\%, and Short-Term Planning (STP) from 53.00\% to 72.30\%. These results empirically validate that explicit, decoupled memory structures are essential for robust, long-horizon nutritional reasoning.

In future work, we plan to extend our framework in two realistic directions. First, we aim to transition from single-agent reasoning to multi-agent collaborative scenarios, where multiple embodied assistants or human-robot teams can share spatial-temporal intelligence to coordinate complex dietary tasks. Second, while Nutri-Vgent currently excels at tracking objective food states and recipe constraints, we plan to incorporate personalized, dynamically updating user profiles (e.g., evolving health goals and taste preferences) directly into the constraint-aware memory, further bridging the gap between academic benchmarks and real-world personalized healthcare.

\section{Conclusion}
\label{sec:conclusion}

In this work, we formalized the novel task of \textit{Embodied Nutrition Management} to bridge continuous visual perception with complex dietary reasoning. To evaluate this, we introduced \textbf{NutriBench-Kitchen}, a benchmark of 1,500 manually verified QA pairs derived from untrimmed cooking videos. Unlike datasets focused on atomic actions, it systematically assesses dynamic inventory tracking, dietary constraints, and long-horizon planning across five dimensions. To address the limitations of general-purpose large vision-language models (LVLMs) in maintaining extended temporal contexts, we proposed \textbf{Nutri-Vgent}. Moving beyond generic retrieval, it features a structured memory architecture that explicitly decouples and synergizes episodic visual observations, persistent food states, and recipe knowledge graphs. Experiments demonstrate that Nutri-Vgent effectively mitigates temporal forgetting and significantly outperforms baselines, particularly in persistent state tracking and constraint satisfaction. These findings underscore that domain-specific, structured memory is essential for robust multimodal reasoning in dynamic kitchen scenes, paving the way for proactive assistants capable of managing daily nutrition.


\bibliographystyle{splncs04}
\bibliography{main}

\end{document}